# Document Similarity from Vector Space Densities

Ilia Rushkin

Harvard University
Cambridge, USA
ilia_rushkin@harvard.edu

**Abstract.** We propose a computationally light method for estimating similarities between text documents, which we call the density similarity (DS) method. The method is based on a word embedding in a high-dimensional Euclidean space and on kernel regression, and takes into account semantic relations among words. We find that the accuracy of this method is virtually the same as that of a state-of-the-art method, while the gain in speed is very substantial. Additionally, we introduce generalized versions of the top-$k$ accuracy metric and of the Jaccard metric of agreement between similarity models.

## 1 Introduction

Estimating similarities among texts is essential for document retrieval, clustering and recommendation systems. Taking into account semantic connections is an interesting way of enhancing the similarity estimation, since it alleviates two interconnected shortcomings of the simpler word-matching methods. One is the difficulty with polysemy and semantics in natural languages, and the other is the near-orthogonality, meaning that the number of matching features is low for most pairs of documents, causing a large uncertainty in the match-based similarities [3]. To incorporate semantic connections, the algorithm should have some kind of a "meaning-distance map" among words. A popular method of doing this is word embedding, also known as vectorization ([5], [7], [9]), where such a map is trained on the frequency of words being near each other in the document corpus. This method achieves mapping words (or, more generally, features such as n-grams) onto points in a Euclidean space in such a way that the Euclidean distance between points represents semantic distance. Word embeddings can now be trained relatively easily, and pre-trained ones are freely available (e.g. [6]).

However, even when the word embedding exists, there remains a question of getting from similarities between features (words) to similarities between documents. At present, it appears that the existing methods for this task are either computationally very heavy (e.g. the relaxed word-moving distance of RWMD, introduced by [8] following [10]), or light but with a significantly lower accuracy (e.g. representing each document by a single centroid point in the embedding space). In this work we propose a method that appears to be both accurate and fast. We call it the density similarity method, or DS for short, since it represents documents as probability densities in the embedding space. On our example data, the DS method performed very similarly to that of RWMD, while being orders of magnitude faster. Furthermore, we think that the idea of the DS method is very natural and straightforward mathematically, giving it interpretability.

The rest of this paper is organized as follows. In Section 2 we introduce and discuss the DS method, with subsections 2.1 and 2.2 devoted to the questions of bandwidth and sampling the space, respectively. In Section 3 we propose new generalizations of accuracy metrics. In particular, they are used in Section 4, where we apply the DS and the benchmark method to several datasets. Section 5 contains the discussion and further thoughts.

## 2   Description of the Method

Word embedding, or vectorization, is a mapping of document features (informally referred to as "words") onto points in a $d$-dimensional Euclidean space $\mathbb{R}^d$, where $d$ is usually in the hundreds. Each document becomes a sequence of points, containing a point multiple times if a word is used multiple times in a document. If we ignore the word order, then the document is a set of distinct points with multiplicity weights assigned to them, i.e. it is a distribution. More formally, the weights form a document-feature matrix $w_{t,i}$ where the index $t = 1, .. N_d$ enumerates documents in the corpus and $i = 1, .. N_f$ enumerates the features. In the simplest case, the elements of this matrix can be the counts of features in documents, but it is common to subject it to various transformations, such as normalizing rows to unit sum and applying the TF-IDF transformation with the goal of lowering weights of those features that are present in too many documents and therefore are not believed to be as informative as the rare features. The word embedding maps $i \to x_i$, where each $x_i$ is a $d$-dimensional vector (the "feature-point"), and therefore a document $t$ is represented by a distribution defined on the set of all feature-points in the corpus: $f_t(x_i) \equiv w_{t,i}$.

The density similarity method that we propose here is to view the document distributions as statistical sample densities and to estimate their distribution density values on a small set of sample points $\{z_j\}$. Transition from the full set of feature-points to the set $\{z_j\}$ involves distances in the embedding space, thus incorporating semantic connections into the method. We call it the "density similarity method".

To estimate the densities, we can use kernel (or "non-parametric") regression, which is a well-developed field (see [4] or [12] for an introduction to it). Namely, let us define the density of a document at an arbitrary point $z \in \mathbb{R}^d$ as a kernel regression:

$$\rho_t(z) = \frac{\sum_{i=1}^{N_f} k\left(\frac{z - x_i}{h}\right) w_{t,i}}{\sum_{i=1}^{N_f} k\left(\frac{z - x_i}{h}\right)} \qquad (1)$$

Here $k(y)$ is the kernel and $h > 0$ is the kernel bandwidth. The most commonly used kernel is Gaussian: $k(y) = (2\pi)^{-d/2} \exp(-|y|^2/2)$. Another notable choice is the radially-symmetric Epanechnikov kernel $k(y) = \max(0, d - |y|^2)$, based on [2], where the $d = 1$ version of this kernel was first studied and demonstrated to be the asymptotically optimal.

We note that the denominator in Eq. 1 is just a normalization constant. For our purposes normalization plays no role, so we could use only the numerator of Eq. 1.

Applying Eq. 1 at the $n$ sample points $z_j$, we obtain the *density matrix* of the corpus: $\rho_{t,j} \equiv \rho_t(z_j)$. Presuming that $n < N_f$, this matrix can be seen as a condensed version of the document-feature matrix $w_{t,i}$, transformed from the $N_f$ original features to $n$ new features. This transformation plays the role similar to latent semantic indexing (LSI) or a topic model such as latent Dirichlet allocation (LDA). Like these examples, the construction of the density matrix incorporates semantic relations among features, since it uses distances in the Euclidean space.

Now the document-document similarity can be extracted as similarity among the rows of the density matrix. Given the interpretation of each row of as a probability distribution, it is conceptually appealing to use the Jensen-Shannon divergence [1], but a simple cosine-similarity measure is a faster alternative.

In the above we were dealing with one corpus of documents. In document retrieval and recommendation problems there are usually two corpora, queries and items, and it is the cross-corpus (query-item) similarity that needs to be found. The generalization of the entire discussion is straightforward: the two matrices $w_{t,i}^{(queries)}$ and $w_{t,i}^{(items)}$ will be independently transformed into two densities $\rho_{t,j}^{(queries)}$ and $\rho_{t,j}^{(items)}$, and similarity will be computed between them.

Eq. 1 is readily generalizable in a variety of ways. For instance, we used a spherically-symmetric kernel, but it is not a requirement. Generally speaking, instead of a single bandwidth parameter $h$ one has a $d \times d$ bandwidth matrix $H$, in which case the kernel arguments $(z - x_i)/h$ in Eq. 1 are replaced by matrix products $H^{-1}(z - x_i)$. However, this level of generality is often too complex, and a common intermediate approximation is to use a diagonal bandwidth matrix $H = \mathrm{diag}(h^{(1)}, \ldots, h^{(d)})$, i.e. each Cartesian axis is served by its own bandwidth. Note, however, that it makes $\rho$ dependent on the choice of the coordinate system in $\mathbb{R}^d$.

Let us briefly discuss the speed of the proposed DS method. If a document contains, on average, $F$ unique features, creation of the density matrix has time complexity of at most $O(FN_d n)$, assuming straightforward calculation of sums in Eq. 1 (although faster methods may exist). The cosine similarity among rows of this matrix (size $N_d \times n$) has time complexity $O(N_d^2 n)$. Hence, the time complexity of the entire calculation is: $O(FN_d n + N_d^2 n)$. By contrast, for the RWMD method the best-case time complexity is $O(N_f^2)$ for each document-document pair ([8]), therefore $O(N_f^2 N_d^2)$ for the entire corpus. The dependence on the number of features (even ignoring the fact that $F < N_f$) is quadratic in RWMD and linear in DS, indicating that on large corpora DS must be significantly faster. In order to make the comparison more direct, let us estimate that $F = N_f N_d^{-b}$. Evidently, $b$ may lie in the interval between two extreme values: 0 (each document contains all the features of the corpus) and 1 (no two documents have any features in common). With this, the time complexities are: $O(N_f N_d^{1-b} n + N_d^2 n)$ for DS, versus $O(N_f^2 N_d^2)$ for RWMD. Below, when we compare RWMD and the DS methods on example datasets, we will see that this indeed translates into a very sizeable difference in speed. We chose RWMD as a benchmark its authors, when introducing it, found it to compare favorably with many other popular algorithms.

The DS method involves a number of meta-parameters:
1. The bandwidth $h$ (or, more generally, the bandwidth matrix $H$).
2. The number $n$ of the sample points.
3. The kernel shape $k(y)$.
4. The type of similarity measure to be applied to the density matrix rows.

In the last two, it is easy to settle on Gaussian kernel and cosine-similarity. These are simple and standard, and the results are not expected to be too sensitive to them. The bandwidth and the sample points, on the other hand, affect the outcome strongly, as is well known for any kernel regression. In principle, the optimal bandwidth and sample points should be found by training on a labeled training set. Such a training, however, is costly, which makes it essential to look for quick estimates of these parameters instead. In the next two sections we discuss our strategy for such estimates.

### 2.1 Choice of Bandwidth

Selection of the bandwidth $h$ in kernel regression like Eq. 1 lies at the heart of kernel regression theory ([4], [12]). The methods fall into two broad categories: computationally heavy methods that perform the optimization of bandwidth (i.e. training), and computationally light "rules-of-thumb". The classic example of the first category is the least-squares cross-validation, which determines $h$ as the minimizer of the following cost function:

$$C(h) = \frac{1}{h^d N^2} \sum_{i,j=1}^{N} \left[ K\left(\frac{x_i - x_j}{h}\right) - \frac{2N}{N-1} k\left(\frac{x_i - x_j}{h}\right) \right] + \frac{2k(0)}{h^d(N-1)} \quad (2)$$

Here $K$ is the convolution kernel: $K(y) = \int k(u)k(y-u)d^d u$. For the Gaussian kernel $k(y) = (2\pi)^{-d/2} \exp(-|y|^2/2)$, it is a broader Gaussian: $K(y) = (4\pi)^{-d/2} \exp(-|y|^2/4)$. In the more general case of a bandwidth matrix $H$, the $h^d$ in the denominators should be replaced with $\det H$, and the kernel arguments $(x_i - x_j)/h$ with $H^{-1}(x_i - x_j)$.

The Silverman rule [11] is an example from the "rule-of-thumb" category. It estimates the bandwidth separately for each Cartesian axis in $\mathbb{R}^d$, i.e. gives a diagonal bandwidth matrix, and a single bandwidth can be formed as the geometric mean of the diagonal. Namely, if $\sigma^{(a)}$ is the standard deviation of the $a$-th components ($a = 1, ..d$) of $x_i$,

$$h_S^{(a)} = \sigma^{(a)} \left(\frac{4}{N(d+2)}\right)^{\frac{1}{d+4}}, \qquad h_S = \left(\prod_{a=1}^{d} h_S^{(a)}\right)^{1/d} \quad (3)$$

Another option is to take the average spacing between the feature points. To estimate the volume populated by the $x_i$, we circumscribe a $d$-dimensional sphere of radius $R$ centered at origin. The log-volume of a ball of radius $R$ is

$$\log v(R) = \log \frac{\pi^{d/2}}{\Gamma(1 + d/2)} + d \log R, \qquad (4)$$

We may also cut out a smaller unpopulated ball of radius $r < R$, so the remaining volume is a spherical layer in which $x_i$ lie. The log-volume of the described spherical layer is

$$\log V(r, R) = \log v(R) + \log\left(1 - e^{\log v(r) - \log v(R)}\right) \qquad (5)$$

In fact, the second logarithm is only a small correction unless $r$ is close to $R$: in a high-dimensional ball, the volume is highly concentrated near the surface. But we keep it here for generality. The bandwidth is then estimated as the typical spacing that the data points would have if they were distributed in that volume uniformly:

$$h_V = e^{(\log V(r,R) - \log N)/d} \qquad (6)$$

The straightforward choices for $r$ and $R$ are the minimum and the maximum of $|x_i|$. However, to avoid making $h_V$ sensitive to outliers, we selected $r$ and $R$ as the 0.1 and 0.9 quantiles of $|x_i|$.

On our data, the volume method of Eq. 6 gave the best results. The Silverman rule gave somewhat smaller bandwidth and led to lower accuracy. The minimization of Eq. 2 was not only computationally heavy, but also produced much smaller bandwidth values, leading to the worst accuracy of all. For this reason, we suggest using Eq. 6 for bandwidth estimation. If labeled training data is available, one can also try to adjust the estimated bandwidth by a multiplicative factor.

### 2.2 Sampling the Space

For a high $d$ (a typical value in word embeddings is 300), creating a regular grid of sample points is not practical due to the "curse of dimensionality": a single cube in $\mathbb{R}^d$ has $2^d$ vertices. The regularity of the sampling array is not required, however. We just need to sample the space sufficiently for measuring distribution differences, wherever they may occur. One solution is to generate any desired number $n$ of sample points randomly, from a distribution that is uniform in a ball of radius $R$. It is essentially the same concept as was considered in the previous section for bandwidth estimation. Logically, one should take a spherical layer $V(r, R)$ rather than a ball, but, as we saw, the difference is negligible unless $r$ is very close to $R$.

As in the previous section, we select $R$ as a high quantile of the norms $|x_i|$ (we used quantile 0.95), rather than $\max|x_i|$, due to the possibility of outliers. The way to generate $n$ uniform random points $z_j$ in a $d$-dimensional ball of radius $R$ is as follows. We first generate $n$ points $Z_j$ from a $d$-variate standard normal distribution (in Cartesian components, it means generating a $d \times n$ matrix of independent standard normal variables). Then, each point is normalized: $z_j = Z_j u_j^{1/d} R / |Z_j|$, where $u_j$ are $n$ numbers independently drawn from the standard uniform distribution on the [0,1] interval.

There is no simple way to choose $n$ optimally. Given the uniform distribution of $z_j$, there is no expectation of results deteriorating if $n$ is too high. Rather, the approximation is always expected to improve with higher $n$, but at an ever-decreasing rate. This situation of "diminishing returns" makes the choice of $n$ similar to that of the number of trees in a random forest algorithm, rather than to the number of components in LSI or the number of topics in LDA. At the same time, since sampling involves randomness, a small $n$ increases the probability of fluctuations in the similarity results. It is even possible for a lower $n$ to give higher accuracy, but not consistently. In our application of the method we tried several values of $n$ ranging from 100 and to 10,000.

## 3      Performance Metrics

On labeled data we can compare the accuracies of the DS method and a benchmark method, and on a non-labeled one we can measure the agreement between them. While standard metrics for accuracy and agreement exist, it seemed useful to us to introduce versions of them that are modified with a "softness" parameter $s$, and present the results for several values of $s$. Of course, we include $s = 0$, at which the modification disappears. The remainder of this section describes these metrics.

The output of a document-similarity method can be described by a distance matrix $M_{t',t}$, where the row-index $t'$ enumerates documents in the corpus of queries and the column-index $t$ – those in the corpus of items. Without loss of generality, we can consider only one query, treating $t'$ as a spectator index. Furthermore, the actual values of distance (equivalently, of similarity) are not important, only their ranks are, because they determine which items, and in what order are returned for a query. Therefore, we consider the vector $r_t$ of ranks of the values in a row of $M$.

An accuracy metric should give more weight to lower ranks: the top-$k$ items are important, but we don't care much in what order it places the almost-irrelevant items in the long tail of high ranks. A standard method is to take only items with rank not higher than a threshold $k$. More generally, even within these items, we can weigh lower ranks higher. Given a labeled data set, we take the $k$ items with the lowest rank, ordered by that rank. The label-correctness values form a 0-or-1 (boolean) vector $c_{k'}$, $k' = 1,..k$, with which we define the *soft top-k accuracy*:

$$A(k,s) = \frac{\sum_{k'=1}^{k} c_{k'} w_{k'}}{\sum_{k'=1}^{k} w_{k'}}, \quad w_{k'} = \frac{1}{k'^s} \tag{7}$$

Here $s \geq 0$ is the softness parameter. The values of $A(k,s)$ lie in the [0,1] interval. At $s = 0$ the metric reduces to the usual top-$k$ average correctness.

Another metric of performance is needed to measure the agreement between two models which produce two ranking vectors, $r_t^{(A)}$ and $r_t^{(B)}$. Choosing a rank threshold $k$, we can measure the agreement as a Jaccard similarity between the items assigned rank $\leq k$ in the two models:

$U^{(A)}(k) = \{t : r_t^{(A)} \leq k\}, \ U^{(B)}(k) = \{t : r_t^{(A)} \leq k\}, \quad U(k) = U^{(A)}(k) \cup U^{(B)}(k),$

with which the Jaccard similarity index is

$$J(k) = \frac{|U^{(A)}(k) \cap U^{(B)}(k)|}{|U(k)|} \tag{8}$$

This metric has a sharp cutoff: if a model ranked an item beyond $k$, it is penalized regardless of *how far beyond $k$* it ranked the item. We want to generalize it to make the penalty depend on the rank. Observe that the quantity $\frac{1}{|U(k)|}\sum_{t \in U(k)} \min\left(1, \left(r_t^{(A)}/k\right)^{-1/s}\right)$ penalizes the model $A$ in just that way (here $s \geq 0$). Taking, for symmetry, the sum of that quantity and its counterpart from model $B$, and subtracting 1, we obtain the *soft Jaccard index*:

$$J(k,s) = -1 + \frac{1}{|U(k)|} \sum_{t \in U(k)} \left[\min\left(1, \left(\frac{r_t^{(A)}}{k}\right)^{-1/s}\right) + [A \to B]\right] \tag{9}$$

By the meaning of $U$, at least one of the two terms in the summand is always 1, so $J(k,s)$ takes values between 0 and 1, and the upper bound is achieved if and only if the two models completely agree on which items they rank within $k$. At $s = 0$, the metric reduces to the standard Jaccard measure of Eq. 8:

$$J(k,0) = -1 + \frac{|U^{(1)}(k)| + |U^{(2)}(k)|}{|U(k)|} = -1 + \frac{|U(k)| + |U^{(1)}(k) \cap U^{(2)}(k)|}{|U(k)|} = J(k)$$

We note in passing that our choice of a power-law dependence was made for simplicity, but one could also generalize $A$ and $J$ using, e.g., an exponential decay.

Recalling now the query-index $t'$, the quantities of Eq. 7 and Eq. 9 can be computed for every query: $A_{t'}(k,s)$, $J_{t'}(k,s)$, and then their distribution over $t'$ can be examined.

## 4    Experiment

Our data set consisted of texts scraped from personal webpages of faculty members and researchers in Harvard University (dataset "People"), the descriptions of events, talks, seminars and public lectures (dataset "Events"), and the news articles recently published at the same institution (dataset "News"). All texts were cleaned and tokenized on the basis of single lemmatized words, dropping stopwords and words shorter than 4 characters. The tokens were then used as features in word embedding. A subset of the dataset "People" was labeled by people's affiliation with Harvard departments (number of classes: 60; class sizes: 10-479, with mean 58 and median 33). Some descriptive statistics of the data are given in Table 1.

**Table 1.** Descriptive statistics of the data.

| Dataset | # documents | Max # tokens/document | Mean # tokens/document |
|---|---|---|---|
| "People" | 11,863 | 66,258 | 1,030 |
| "People" labeled | 3,495 | 55,565 | 1,696 |
| "Events" | 5,177 | 535 | 30 |
| "News" | 5,118 | 2,115 | 146 |

The "People" dataset always plays the role of the "queries" corpus, and the role of the "items" corpus can be played by any of the three datasets (when the same dataset "People" is used as "queries" and as "items", we remove self-recommendations).

We used the same word2vec embedding for all methods, pre-trained on a corpus of Wikipedia articles, with dimensionality $d = 300$ ([6]). The tokens that are absent in our corpora were dropped from the embedding, leaving 138,030 tokens.

We computed the query-item similarity matrices using the RWMD method and using the DS method with several different numbers of sample points. As was mentioned earlier, the RWMD model for document similarity was shown by its authors to compare favorably with a number of other algorithms, which is why we confine ourselves to using it as a benchmark. All calculations were done on the same machine (2.4 GHz processor and 16 GB memory), in an R environment. For RWMD, we used the implementation of this method in the R package *text2vec*, with Euclidean distance. Creation of the density matrices was done with a custom R script, and the subsequent calculation of the similarity matrices was done using the R package *quanteda*.

In the DS method, the kernel was Gaussian, and the bandwidth was determined by Eq. 6, with several adjustment factors tried from 0.25 to 64, in powers of 2. We first used the dataset "People" as both queries and items (recommendation of people to other people), and looked at the top-5 accuracy on the labeled subset of this dataset. As Fig. 1 illustrates, the accuracy of RWMD is only slightly higher than that of density similarity with 500 or more points. For 1,000 and 10,000 points the difference from RWMD is statistically insignificant ($p > 0.1$).

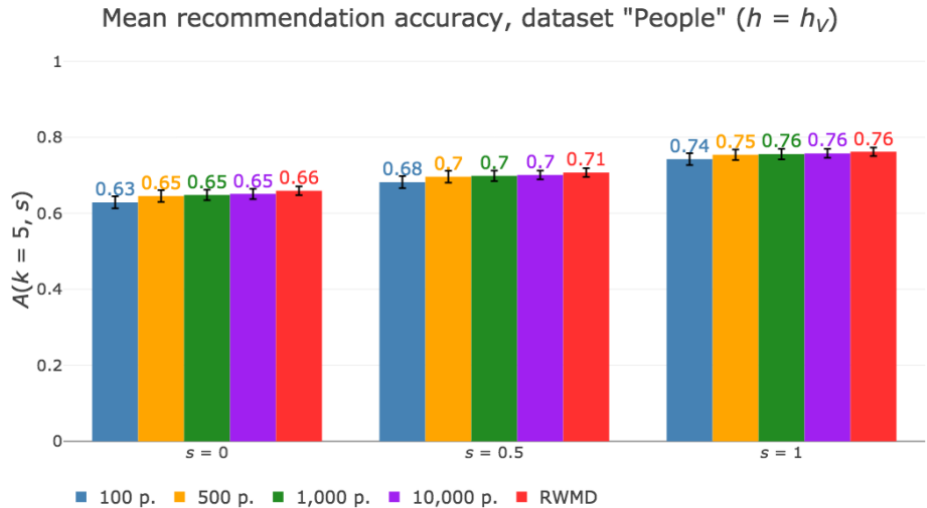

**Fig. 1.** Top-5 accuracy comparison among RWMD and several versions of density similarity method, differing by the number of sample points. Bandwidth is $h_V$ from Eq. 6. The vertical axis is the accuracy measure of Eq. 7, averaged across queries. The horizontal axis is the softness parameter. The error-bars show 3 standard errors.

Larger rank-cutoffs demonstrate a similar picture: top-10 accuracy values are consistently lower than the top-5 values, but the difference between RWMD and DS with 1,000 or 10,000 points not exceeding 0.02. At the same time, the difference in computation speed was large: the RWMD calculation took over 100 hours, whereas the density similarity with 1,000 points took less than 11 minutes (more precisely, 654 sec., consisting of 392 sec. on density estimation and 262 sec. on the similarity matrix), and the calculation with 500 points took about 5 minutes (175 sec. on density estimation and 124 sec. on the similarity matrix).[1]

We also calculated the recommendations of news and events to people, where the dataset "People" serves as queries and is paired either with "Events" or "News" as items. The RWMD method for both recommendations together took about 60,000 sec. (about 16,000 sec. for "News" and 45,000 sec. for "Events"). By comparison, the density similarity method with 1,000 sample points took about 930 seconds for the same task, and with 500 sample points – under 600 sec.

We can measure the agreement with RWMD on unlabeled datasets using the soft Jaccard index of Eq. 9, and the results are shown in Fig. 3. They show a strong overall correlation between the model outcomes (this gives no indication which model gives a better recommendation when they do not agree).

As a minimal form of training the model, we repeated the calculation using adjusted bandwidth: multiplying the Eq. 6 by several simple factors. This exploration showed that $h_V$ is close to the optimum. Significant deviations from it lead to a decrease either in accuracy, or in the Jaccard index of agreement with RWMD, or in both. For instance, Fig. 2 shows the accuracy at a halved bandwidth which is lower than in Fig 1.

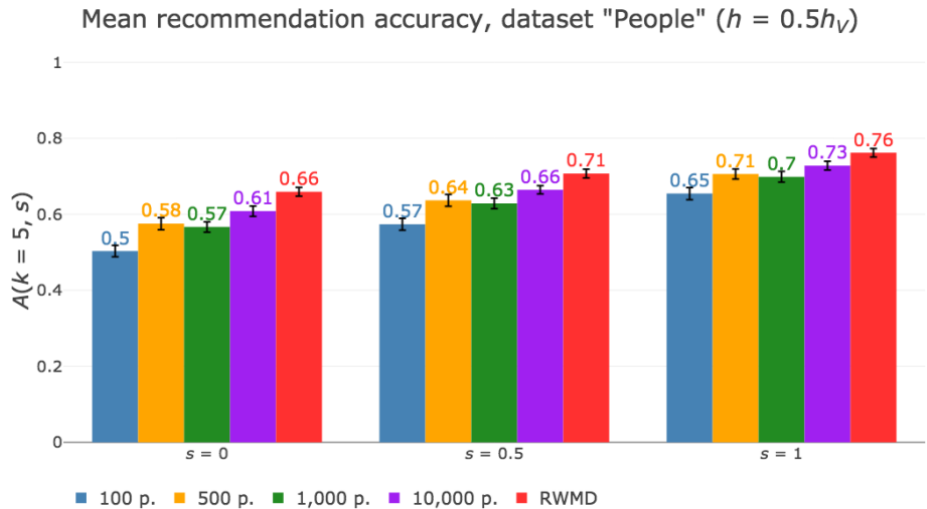

**Fig. 2.** Accuracy comparison among RWMD and several versions of density similarity method, differing by the number of sample points. Bandwidth is adjusted by a factor 0.5. The vertical axis is the accuracy measure of Eq. 7, averaged across queries. The horizontal axis is the softness parameter. The error-bars show 3 standard errors.

---

[1] In this case, queries and items are represented by the same corpus, so only one density matrix is needed.

## 5   Discussion and further work

The density similarity (DS) method, which we propose here, estimates document similarities – a crucial task for document retrieval and clustering.

The speed of the DS method strongly depends on the number of sample points. We found that a sample of 500 or 1,000 points is sufficient: increasing it further produces only a small additional improvement. Even with 10,000 sample points, the DS method is much faster than RWMD, while its top-$k$ accuracy turns out essentially the same. We believe that the gain in speed compensates well for the slight difference in accuracy, even if it turns out that that difference is systematic. Elsewhere, RWMD has been shown to be more accurate than a number of other popular methods ([8]), and by amounts that are significantly larger than the difference that we observe here.

Our application of the DS method relies on direct estimates of meta-parameters (bandwidth, sampling the space). In this form, it is an unsupervised machine learning algorithm. However, if a labeled dataset is available, it is straightforward to incorporate some training into the method, as we did with the bandwidth adjustment coefficient. We found the bandwidth is the single most important parameter of the method – as is typical in non-parametric regressions.

The DS method is essentially a kernel regression in the embedding space. In our view, it is a very straightforward idea, making the results easier to interpret and the method – easier to develop further. Moreover, the corpus density matrix, computed as a step of the method, is an interesting condensed version of the document-feature matrix, and can be used as such for purposes other than finding document similarity, e.g. clustering, or visual representations of document corpora.

In the future, we hope to pursue several possible directions of further research: generalization of document features from single words to n-grams; sensitivity to transformations of the document-feature matrices (in this work we used a standard TF-IDF transformation); possibility of combining this method with others in a multi-step fashion.

## 6   Acknowledgments

The author is grateful for the support from the Office of the Vice Provost for Advances in Learning at Harvard University.

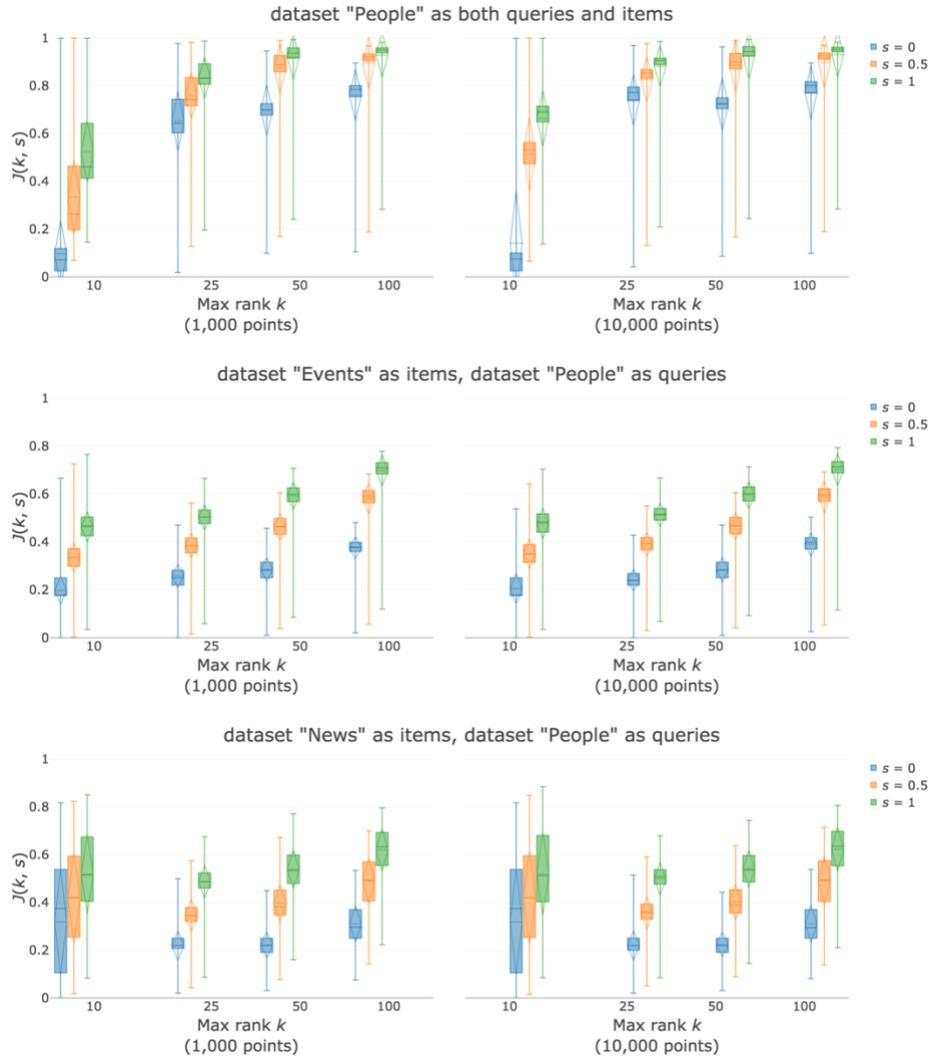

**Fig. 3.** Agreement of two versions of density similarity methods (differing in the number of sample points) with RWMD in three recommendation problems: recommending people to people, events to people, and news to people. The vertical axis is the generalized Jaccard measure of Eq. 9. The colors represent three values of the softness parameter. The box plots show the quartiles of the distribution of this quantity across all queries, and the added diamond shapes show the mean values $\pm$ one standard deviation.